\documentclass[11pt]{article}
\usepackage[final]{acl}

\usepackage{times}
\usepackage{latexsym}
\usepackage[T1]{fontenc}
\usepackage[utf8]{inputenc}
\usepackage{microtype}
\usepackage{inconsolata}

\usepackage{graphicx}
\usepackage{amsmath}
\usepackage{amssymb}
\usepackage{amsfonts}
\usepackage{booktabs}
\usepackage{multirow}
\usepackage{kotex}

\title{Phrase-Localized Language-Contrastive Guidance: Training-Free Localized Accent Control for Code-Switching Text-to-Speech}

\author{
\textbf{Che Hyun Lee\textsuperscript{1}} \quad 
\textbf{Sangkwon Park\textsuperscript{1}} \quad
\textbf{Donghun Kang\textsuperscript{1}} \quad
\textbf{Dongwook Lee\textsuperscript{2}} 
\\
\textbf{Youngho Cho\textsuperscript{2}} \
\textbf{Heeseung Kim\textsuperscript{3}}\footnotemark[2] \
\textbf{Sungroh Yoon\textsuperscript{1,2,4}}\footnotemark[2]
\\
\footnotetext[2]{Corresponding authors.}
\\
\textsuperscript{1}Department of Electrical and Computer Engineering, Seoul National University \\
\textsuperscript{2}Interdisciplinary Program in Artificial Intelligence, Seoul National University \\
\textsuperscript{3}Department of Artificial Intelligence, University of Seoul \\ \textsuperscript{4}AIIS, ASRI, INMC, and ISRC, Seoul National University
}

\begin{document}
\maketitle

\renewcommand{\thefootnote}{\fnsymbol{footnote}} 
\footnotetext[2]{Corresponding authors.}

\begin{abstract}
Current speech synthesis struggles with code-switching, which mixes a foreign language phrase into a primary language utterance, causing the phrase to be spoken with the primary language's accent rather than its native one. We propose Phrase-Localized Language-Contrastive Guidance (LCG), a training-free inference framework that restores a native accent to code-switched phrases in cross-lingual text-to-speech. LCG replaces the single language guidance applied across the whole utterance with a separate guidance for each region, so each part is guided by its own language. To choose where to apply this localized guidance, we propose a self-attention probing technique that finds the phrase boundaries without external alignments. Together, these components generate speech in which each region carries the accent of its own language, requiring no fine-tuning or auxiliary models. Across diverse language pairs, LCG robustly increases the nativeness of the code-switched phrase while suppressing accent leakage, and preserving overall speaker identity and naturalness.\footnote{https://saga1214.github.io/PhraseLocalizedLCG/}
\end{abstract}

\section{Introduction}
\label{sec:intro}

The emergence of zero-shot multilingual text-to-speech (TTS) foundation models has enabled high-fidelity cross-lingual voice cloning without task-specific fine-tuning~\citep{anastassiou2024seedttsfamilyhighqualityversatile,casanova24_interspeech,du2024cosyvoice,wang2025maskgct,zhang2023speak,zhu2026omnivoice}. Trained on massive multilingual datasets, these models successfully transfer a speaker's identity to unseen languages. However, despite the remarkable capacity, they still suffer from severe \emph{cross-lingual accent leakage} resulting in an unnatural foreign accent when the language of the reference voice prompt does not match the target text.

This leakage becomes acute in intra-utterance code-switching (CS)~\citep{sitaram2020surveycodeswitchedspeechlanguage}, where a primary language (\emph{matrix carrier}) and a foreign insertion (\emph{embedded phrase}) mix within a single sentence~\citep{10.1093/oso/9780198240594.001.0001}. A robust CS TTS system must maintain carrier fluency and speaker consistency while pronouncing the foreign phrase with a native accent~\citep{9053094,9362099}. This is inherently challenging because the speaker's voice profile locks the model into a primary language, making it difficult to shift pronunciation styles for just a specific, localized segment.

\begin{figure*}[t]
\centering
\includegraphics[width=1.0\linewidth]{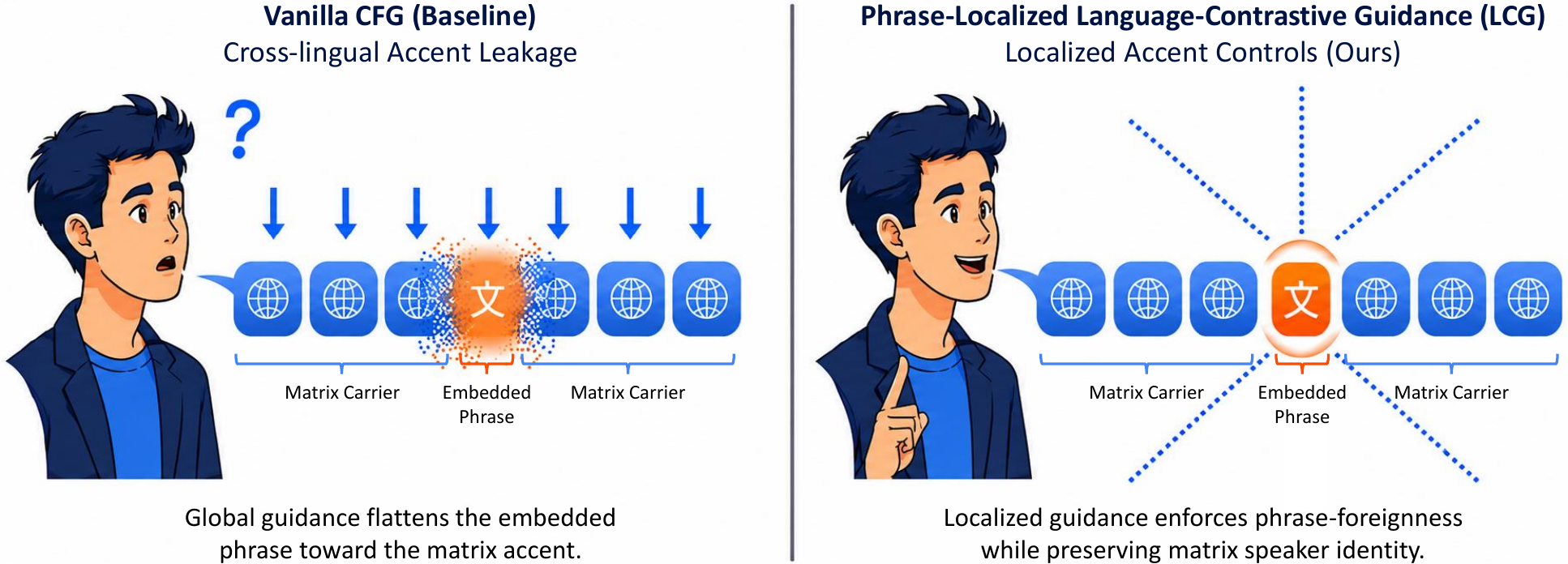}
\caption{\textbf{Conceptual illustration of cross-lingual accent leakage and localized control.} (Left) The unguided \textbf{Baseline} flattens the foreign embedded phrase into the matrix carrier language accent, causing mispronunciation and acoustic distortion around the embedded segment. (Right) Our proposed \textbf{LCG} framework applies localized guidance over the self-derived phrase boundaries, enforcing authentic native pronunciation while preserving the surrounding matrix carrier quality and speaker identity.}
\label{fig:cs_illustration}
\vskip -0.2in
\end{figure*}

This accent bottleneck stems from two interconnected issues. First, zero-shot voice cloning inherently entangles speaker identity with the prompt's native language habits, making it difficult to separate voice characteristics from foreign pronunciation. Second, while Classifier-Free Guidance (CFG)~\citep{ho2021classifierfree} is widely adopted to improve text adherence and audio quality, it heavily amplifies this reference voice bias globally. Because standard CFG applies a uniform scale across the entire sequence, it cannot apply stronger or weaker guidance to specific segments~\citep{10657946}. Instead, as in Fig.~\ref{fig:cs_illustration}, it overrides local language transitions, aggressively flattening the embedded phrase's accent into the matrix carrier.

Notably, this accent leakage persists even in massive multilingual models like OmniVoice~\citep{zhu2026omnivoice}, a discrete diffusion language model (DLM)-based TTS model that supports over 600 languages. To resolve this bottleneck, we introduce a training-free, inference-time localized guidance framework. Unlike prior methods that apply CFG only to predetermined, selected tokens~\citep{zheng2026selectiveclassifierfreeguidancezeroshot}, our framework operates on frame-level span masks dynamically derived from internal self-attention probing to achieve precise acoustic control.
Building on this, we propose \textbf{Phrase-Localized Language-Contrastive Guidance (LCG)}. LCG transposes contrastive steering onto localized audio sequences by swapping language tags over the refined boundaries. This localized control integrates seamlessly with the bi-directional denoising process of discrete diffusion models, which naturally absorbs minor mask spillovers while keeping the matrix carrier and speaker identity intact.

In summary, our main contributions are threefold: (1) we formalize the \textbf{Phrase-Localized LCG} framework, which mathematically decouples the language-steering scale from global guidance to enable independent accent control, (2) we introduce a training- and module-free phrase localization method via \textbf{internal attention probing}, showing that an expanded recall masking strategy aligns naturally with DLMs to absorb minor boundary errors while keeping the carrier clean, and (3) we construct and open-source a balanced, \textbf{1,200-utterance synthetic code-switching benchmark corpus} spanning 12 directions across 5 languages to facilitate reproducible research. Extensive evaluations show that LCG robustly enforces native accents across all directions, maximizing foreign accent clarity while preserving global naturalness.
\section{Related Works}
\label{sec:related}

\begin{figure*}[t]
\centering
\includegraphics[width=1.0\linewidth]{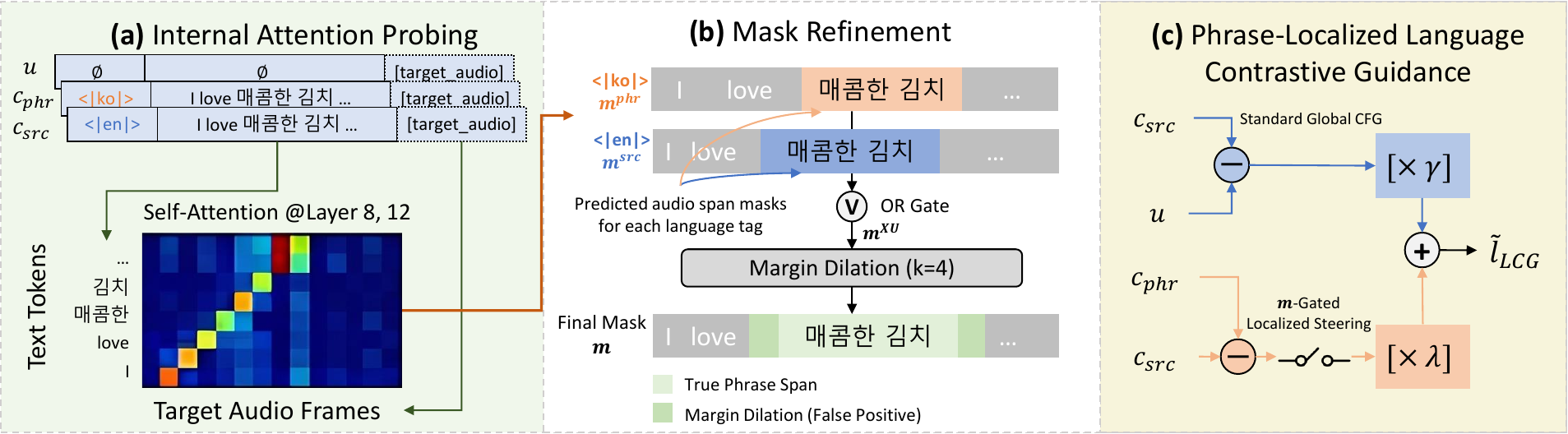}
\caption{\textbf{Overall architecture of the proposed training-free localized guidance framework.} 
An intra-utterance code-switching script mixing an English matrix carrier (``I love'') with an embedded Korean phrase (``매콤한 김치'', meaning ``spicy kimchi'') illustrates the pipeline.
\textbf{(a) Internal Attention Probing:} Audio-to-text frame alignments are extracted from decoder layers ($L_8, L_{12}$) under language-conditioned source ($m^{\text{src}}$) and foreign phrase ($m^{\text{phr}}$) prefixes to derive the raw phrase mask $m$. 
\textbf{(b) Mask Refinement:} Resolves alignment shifts via a dual-tag mask union ($m^{\text{src}} \lor m^{\text{phr}}$) followed by a symmetric margin dilation ($k=4$) to maximize boundary recall. 
\textbf{(c) Phrase-Localized Language-Contrastive Guidance:} Synthesizes the final logits $\tilde{l}_{\text{LCG}}$ by utilizing $m$ as a localized gating switch, decoupling the independent language-steering scale ($\lambda$) from the global text guidance scale ($\gamma$).}
\label{fig:method_overview}
\vskip -0.2in
\end{figure*}

\paragraph{Multilingual and Code-Switching Speech Synthesis.}
Recently, multilingual foundation models clone speaker identities remarkably well across unseen single-language tracks \citep{10842513, casanova24_interspeech,du2024cosyvoice,fan2026lladattsunifyingspeechsynthesis,peng-etal-2024-voicecraft,wang2025maskgct,zhang2023speak,zhu2026omnivoice}. However, they frequently struggle when multiple languages mix dynamically within a single sentence \citep{sitaram2020surveycodeswitchedspeechlanguage}. Traditional code-switching TTS (CS TTS) approaches address this by baking capabilities directly into model weights via bilingual posteriorgrams \citep{9053094}, cross-lingual embeddings \citep{9362099}, multi-stage synthetic fine-tuning \citep{Xu2024EnhancingCT}, or specialized diffusion architectures \citep{cho22_interspeech,kim24h_interspeech,10687605,10889531,10891773,yang24i_interspeech,11462369}. More recently, contemporary large-scale frameworks such as X-Voice \citep{xu2026xvoiceenablingspeak30} attempt to alleviate accent leakage via dual-level language injection architectures; however, these methods rely on massive multilingual corpus fine-tuning and, as explicitly acknowledged in \citet{xu2026xvoiceenablingspeak30}, still struggle to optimize intra-sentential code-switching constructs. Conversely, our framework is entirely inference-only and training-free, evaluated on a balanced 1,200-utterance corpus that specifically targets dense multi-word phrasal insertions rather than the unmanaged or single-word alternations dominant in prior zero-shot benchmarks~\citep{lyu10_interspeech,paik-etal-2026-hike,xie2026switchlingua,ugan2025pier}.

\paragraph{Inference-Time Control and Spatial CFG.}
Classifier-Free Guidance (CFG)~\citep{ho2021classifierfree} is the standard test-time knob for diffusion-based speech generation~\citep{ju2024naturalspeech,le2023voicebox,pmlr-v202-liu23f}, and has recently been extended to discrete language model decoding~\citep{sanchez2023stay,zheng2026selectiveclassifierfreeguidancezeroshot}. Crucially, however, standard speech-side frameworks apply the guidance scalar uniformly across the entire sequence, lacking the spatial resolution to manage localized phonetic shifts. While contemporary scaling efforts like X-Voice \citep{xu2026xvoiceenablingspeak30} introduce decoupled CFG paths, their optimization operates globally across the \emph{temporal generation axis} (timesteps) via decay schedules, failing to isolate localized structural boundaries. While region-conditioned steering is widely utilized in text-to-image spatial masks~\citep{liu2022compositional,10657946,Brooks_2023_CVPR,Zhang_2023_ICCV}, no prior work brings a localized, frame-level spatial guidance paradigm into discrete non-autoregressive speech LMs~\citep{wang2025maskgct,zhu2026omnivoice}. Our proposed Phrase-Localized LCG framework fills this gap with zero training cost, deriving tracking masks dynamically from internal self-attention paths while executing contrastive cross-lingual tag swaps natively at runtime.

\paragraph{Attention Probing and Sequence Alignment.}
Attention in TTS traditionally serves to align text tokens with acoustic frames, evolving from soft matrices \citep{wang17n_interspeech,8461368} to hard monotonic searches \citep{NEURIPS2020_5c3b99e8,pmlr-v139-kim21f} and forced-alignment pipelines \citep{mcauliffe17_interspeech}. Concurrently, attention probing in language modeling remains a passive, post-hoc diagnostic tool. Our work departs from this setup by closing the loop between probing and dynamic sequence conditioning. Instead of relying on external tools, we dynamically extract implicit alignments from the DLM's internal self-attention layers during the forward pass to establish a localized phrase mask. Crucially, the iterative, bi-directional denoising process of discrete diffusion samplers naturally acts as a buffer that absorbs minor boundary mask errors, ensuring rigorous phrase-foreignness without inducing acoustic or identity discontinuities.
\section{Method}
\label{sec:method}

Our training-free inference framework controls cross-lingual accents by guiding the logits of a pretrained discrete diffusion language model (DLM). As illustrated in the macro pipeline in Fig.~\ref{fig:method_overview}, the operational layout is split into three sequential stages: (a) frame-level mask extraction via self-attention probing (Sec.~\ref{ssec:span}), (b) asymmetric boundary expansion to optimize recall (Sec.~\ref{ssec:refine}), and (c) localized scale adjustment via independent language-contrastive steering (Sec.~\ref{ssec:lcg}).

\subsection{Problem Statement \& Preliminaries}
\label{ssec:preliminaries}

OmniVoice \citep{zhu2026omnivoice} is a discrete DLM-based text-to-speech (TTS) framework capable of synthesizing over 600 languages. Operating over an audio codec vocabulary, it iteratively unmasks an acoustic sequence over $T=32$ steps. At each generation step, the model input is configured as the concatenation of control and data streams:
\begin{equation}
\label{eq:prefix}
\underbrace{[\text{lang\_tag}]}_{\text{1 token}}\;\underbrace{[\text{transcript}]}_{\text{text tokens}}\;\underbrace{[\text{ref\_audio}]}_{\text{voice prompt}}\;\underbrace{[\text{target\_audio}]}_{\text{partially masked}}
\end{equation}
where auxiliary control tokens and sequence delimiters are omitted for brevity \citep{zhu2026omnivoice}. Based on this prefix, the model predicts vocabulary logits for the masked regions. To maximize speech naturalness and text adherence, OmniVoice utilizes Classifier-Free Guidance (CFG) \citep{ho2021classifierfree} via a two-pass conditional and unconditional inference step. Standard CFG at scale $\gamma$ combines these logits as:
\begin{equation}
\tilde{l} \;=\; c_{\text{src}} \;+\; \gamma\,(c_{\text{src}} - u),
\label{eq:cfg}
\end{equation}
where $c_{\text{src}}$ represents the primary language-conditioned logit and $u$ denotes the unconditional logit where the prefix tokens are entirely dropped.

However, pretrained cross-lingual frameworks inherently suffer from cross-lingual accent leakage during intra-utterance code-switching (CS) setups comprising a \emph{matrix carrier} (primary language) and a foreign \emph{embedded phrase}. When conditioned on a single global language tag, the standard guidance scale $\gamma$ in Eq.~\ref{eq:cfg} uniformly biases every acoustic frame toward $c_{\text{src}}$, aggressively flattening the embedded phrase's accent into the matrix carrier and collapsing localized phrase-foreignness. To circumvent this limitation without fine-tuning, we introduce a third, parallel \emph{phrase language-conditioned logit} ($c_{\text{phr}}$) that swaps the input language tag to match the embedded phrase, enabling independent accent steering. The full algebraic integration of $c_{\text{phr}}$ is detailed in Sec.~\ref{ssec:lcg}.

\subsection{Phrase Localization via Self-Attention}
\label{ssec:span}

To apply localized foreignness control, we dynamically extract a per-frame binary mask $m \in \{0,1\}^{N_{\text{frames}}}$ over the target acoustic sequence during the forward pass, bypassing the need for auxiliary alignment models (Fig.~\ref{fig:method_overview}(a)). Specifically, for each generated acoustic frame $a$, we monitor the self-attention weights within the decoder layers and isolate the text token index receiving the maximum weight via an argmax operation. If this argmax token falls within the character span of the embedded phrase, we set $m[a] = 1$; otherwise, $m[a] = 0$.

\begin{figure}[t]
\centering
\includegraphics[width=0.95\linewidth]{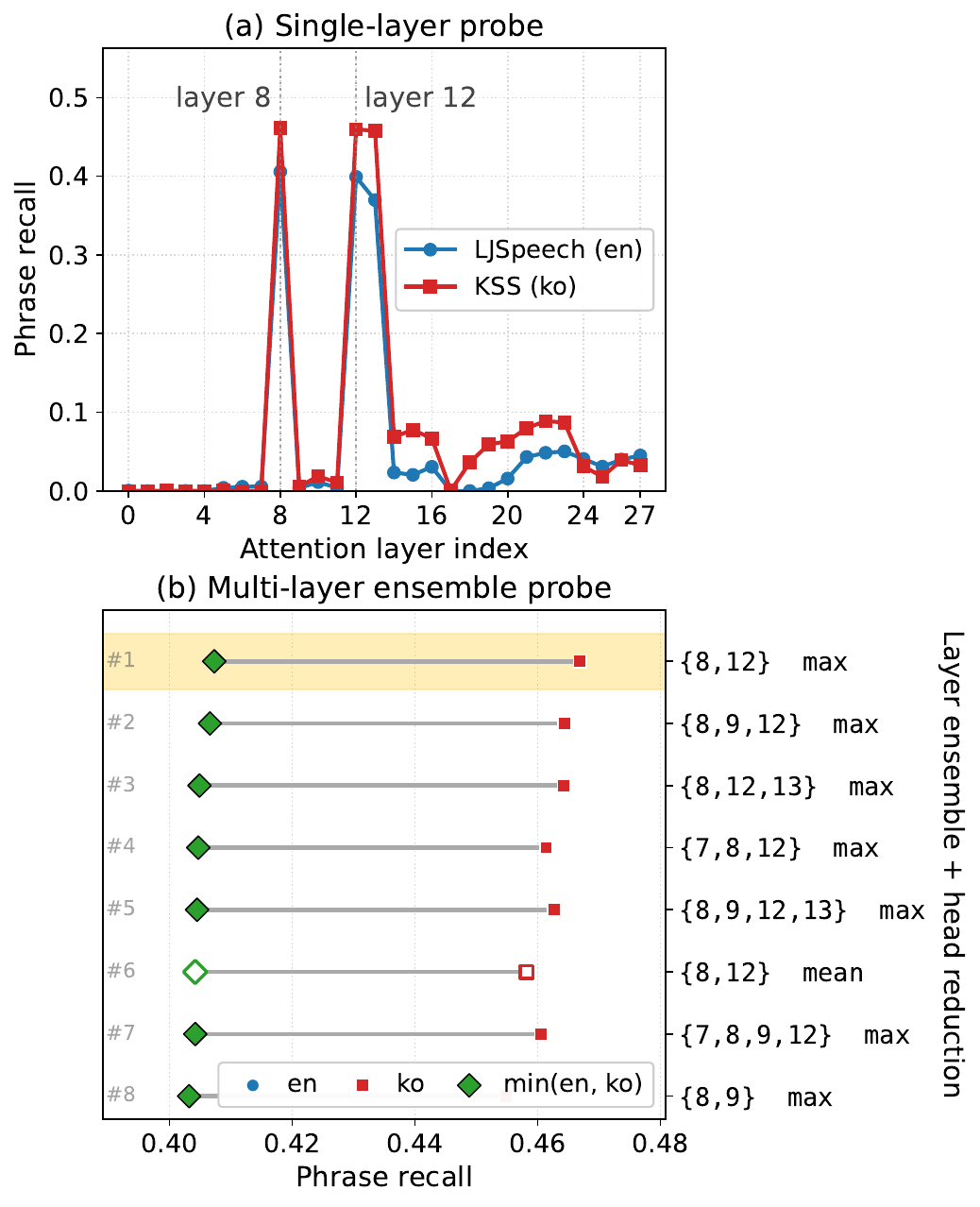}
\caption{\textbf{Attention probe exploration} on the \texttt{KSS} and \texttt{LJSpeech} pilot sets. \textbf{(a) Single-layer probe:} Recall across 28 decoder layers, showing sharp, language-agnostic alignment tracking spikes at $L_8$ and $L_{12}$. \textbf{(b) Multi-layer ensemble:} Comparison of top configurations and head reductions. The Rank-1 setup ($\{L_8, L_{12}\}$ max, highlighted in yellow) maximizes worst-case boundary recall, denoted by the rightmost joint minimum ($\min(\text{en}, \text{ko})$, green diamond).}
\label{fig:attention_probe}
\vskip -0.2in
\end{figure}

To determine the optimal layers for this tracking, we analyze a typologically distinct pilot set comprising Korean (\texttt{KSS})~\cite{park2018kss} and English (\texttt{LJSpeech})~\cite{ljspeech17} utterances, which mitigates script-specific bias and isolates domain-agnostic alignment traits. As illustrated in Fig.~\ref{fig:attention_probe}, our empirical layer-wise probing reveals two critical network properties: alignment tracking spikes sharply and exclusively within the mid-decoder layers (Fig.~\ref{fig:attention_probe}(a)), and a multi-layer ensemble of $\{L_8, L_{12}\}$ combined with head-wise max-pooling consistently maximizes the worst-case boundary recall across both domains (Fig.~\ref{fig:attention_probe}(b)).

\paragraph{Asymmetric Alignment Error Tolerance.}
The selection of our attention configuration is governed by a stark precision-recall asymmetry inherent to localized steering. A false negative (recall loss) is catastrophic: any missed frame within the embedded phrase region defaults to the matrix accent, triggering an irreversible phrase-foreignness collapse. Conversely, a false positive (precision loss) is benign, as accidental steering over the matrix carrier region is easily overridden by the dominant textual context of the surrounding matrix carrier language tokens. Furthermore, since self-attention patterns dynamically shift across generation timesteps, any transient alignment errors are naturally smoothed out and corrected over the course of the iterative denoising process. This asymmetric operational tolerance justifies prioritizing inclusive boundary coverage over tight precision, directly motivating the explicit mask refinement strategies in Sec.~\ref{ssec:refine}.

\subsection{Mask Refinement}
\label{ssec:refine}

While the raw argmax mask $m^{\text{src}}$, obtained by applying primary language-conditioning to the mask $m$ from Sec.~\ref{ssec:span}, is highly precise near the center of the embedded phrase, it tends to be conservative at the boundaries. Based on the asymmetric alignment error tolerance (Sec.~\ref{ssec:span}), we introduce two zero-overhead refinement strategies to aggressively maximize mask recall (Fig.~\ref{fig:method_overview}(b)): we sequence these operations by first computing a dual-tag union and subsequently applying a margin dilation.

\paragraph{Dual-Tag Mask Union.} During inference, our framework computes self-attention tensors for multiple conditional logits at once. The standard matrix-conditioned logit yields the mask $m^{\text{src}}$ using the original matrix carrier tag, whereas the phrase language-conditioned logit substitutes it with the embedded phrase tag to compute an alternative mask $m_{\text{phr}}$ via the same argmax procedure in Sec.~\ref{ssec:span}. Because switching the global language token slightly shifts the self-attention alignment tracks at the boundaries, we merge these complementary profiles via an elementwise logical union:
\begin{equation}
\label{eq:union}
m^{\text{XU}} \;=\; m^{\text{src}} \;\lor\; m^{\text{phr}}.
\end{equation}
This union effectively captures boundary frames missed by either individual tag path. Since both attention tensors are already computed during the parallel conditional forward logits, this refinement introduces no additional computational overhead.

\paragraph{Margin Dilation.} To account for residual boundary frames after union, we subsequently apply a symmetric graded dilation to the union mask. For $s = 1, \dots, k$ we expand the current mask to its $\pm s$-frame neighborhood, so that the successive radii accumulate into a single max-filter of triangular radius:
\begin{equation}
\begin{aligned}
m[a] &= \max_{|\delta| \le r_k} m^{\mathrm{XU}}[a + \delta], \\
r_k  &= \sum_{s=1}^{k} s
     = \frac{k(k+1)}{2}.
\end{aligned}
\end{equation}
Empirically, $k = 4$, an effective $\pm 10$-frame window, maximizes boundary coverage, denoted as \texttt{M4+XU}. As in Fig.~\ref{fig:method_overview}(b), the over-expanded false positive zones resulting from this dilation inevitably spill over into the surrounding matrix carrier region. However, these transient boundary errors are safely insulated and naturally neutralized within the diffusion model's iterative denoising buffer, preserving global matrix carrier naturalness while unlocking unconstrained local foreignness.

\subsection{Language-Contrastive Guidance}
\label{ssec:lcg}

\begin{table*}[t]
\centering
\footnotesize
\setlength{\tabcolsep}{4.8pt}
\begin{tabular}{l c c c c c c c c c c c c}
\toprule
 & \multicolumn{6}{c}{Linguistic Accuracy \& Accent Pronunciation} & \multicolumn{6}{c}{Acoustic Quality \& Speaker Consistency} \\
\cmidrule(lr){2-7} \cmidrule(lr){8-13}
 & \multicolumn{2}{c}{$\mathrm{MER}\downarrow$} & \multicolumn{2}{c}{$\mathrm{LA}_e\uparrow$} & \multicolumn{2}{c}{$\mathrm{LID}_e\uparrow$} & \multicolumn{2}{c}{UTMOS$\uparrow$} & \multicolumn{2}{c}{SIM$\uparrow$} & \multicolumn{2}{c}{$\mathrm{SIM}_{m\leftrightarrow e}\uparrow$} \\
\cmidrule(lr){2-3} \cmidrule(lr){4-5} \cmidrule(lr){6-7} \cmidrule(lr){8-9} \cmidrule(lr){10-11} \cmidrule(lr){12-13}
Language Pair & base & ours & base & ours & base & ours & base & ours & base & ours & base & ours \\
\midrule
$\text{EN}\to\text{JA}$   & 0.623 & \textbf{0.403} & 0.000 & \textbf{0.580} & 0.043 & \textbf{0.900} & 4.450 & 4.060 & 0.960 & 0.941 & 0.895 & 0.890 \\
$\text{JA}\to\text{EN}$   & 0.312 & \textbf{0.233} & 0.816 & \textbf{0.906} & 0.664 & \textbf{0.807} & 3.270 & 3.280 & 0.977 & 0.973 & 0.977 & 0.977 \\
\addlinespace
$\text{EN}\to\text{KO}$   & 0.337 & \textbf{0.111} & 0.494 & \textbf{0.887} & 0.585 & \textbf{0.993} & 4.370 & 3.710 & 0.949 & 0.926 & 0.842 & 0.833 \\
$\text{KO}\to\text{EN}$   & 0.385 & \textbf{0.371} & 0.592 & \textbf{0.718} & 0.355 & \textbf{0.486} & 3.290 & 3.340 & 0.976 & 0.975 & 0.974 & 0.972 \\
\addlinespace
$\text{DE}\to\text{JA}$ & 0.746 & \textbf{0.460} & 0.011 & \textbf{0.485} & 0.073 & \textbf{0.632} & 3.570 & 3.520 & 0.975 & 0.970 & 0.979 & 0.975 \\
$\text{JA}\to\text{DE}$ & 0.742 & \textbf{0.655} & 0.333 & \textbf{0.644} & 0.423 & \textbf{0.712} & 3.070 & 2.990 & 0.974 & 0.972 & 0.980 & 0.982 \\
\addlinespace
$\text{DE}\to\text{KO}$ & 0.525 & \textbf{0.320} & 0.029 & \textbf{0.578} & 0.073 & \textbf{0.721} & 3.530 & 3.400 & 0.979 & 0.977 & 0.983 & 0.980 \\
$\text{KO}\to\text{DE}$ & 0.555 & \textbf{0.499} & 0.182 & \textbf{0.350} & 0.188 & \textbf{0.390} & 3.210 & 3.160 & 0.975 & 0.975 & 0.976 & 0.975 \\
\addlinespace
$\text{FR}\to\text{JA}$ & 0.747 & \textbf{0.607} & 0.005 & \textbf{0.130} & 0.026 & \textbf{0.191} & 2.960 & 2.890 & 0.985 & 0.982 & 0.988 & 0.986 \\
$\text{JA}\to\text{FR}$ & 0.683 & \textbf{0.644} & 0.203 & \textbf{0.468} & 0.378 & \textbf{0.687} & 3.060 & 3.050 & 0.972 & 0.974 & 0.980 & 0.981 \\
\addlinespace
$\text{FR}\to\text{KO}$ & 0.534 & \textbf{0.460} & 0.111 & \textbf{0.409} & 0.140 & \textbf{0.473} & 2.890 & 2.830 & 0.984 & 0.983 & 0.987 & 0.987 \\
$\text{KO}\to\text{FR}$ & 0.576 & \textbf{0.572} & 0.022 & \textbf{0.064} & 0.017 & \textbf{0.063} & 3.260 & 3.190 & 0.975 & 0.976 & 0.973 & 0.975 \\
\midrule
\midrule
\textbf{12-dir Overall} & \textbf{0.564} & \textbf{0.445} & \textbf{0.233} & \textbf{0.518} & \textbf{0.247} & \textbf{0.588} & \textbf{3.411} & \textbf{3.285} & \textbf{0.973} & \textbf{0.969} & \textbf{0.961} & \textbf{0.959} \\
$\Delta$ (Ours $-$ Base) \quad  & \multicolumn{2}{c}{\textbf{$-0.119$}} & \multicolumn{2}{c}{\textbf{$+0.285$}} & \multicolumn{2}{c}{\textbf{$+0.341$}} & \multicolumn{2}{c}{\textbf{$-0.126$}} & \multicolumn{2}{c}{\textbf{$-0.004$}} & \multicolumn{2}{c}{\textbf{$-0.002$}} \\
\bottomrule
\end{tabular}
\caption{\textbf{Comprehensive Headline Comparison.} Performance of the vanilla unguided baseline (\texttt{base}) versus our final configuration (\texttt{M4+XU}, $\lambda=7$) across all twelve code-switching directions. The final block summarizes the macro-average performance across the entire 12-direction evaluation corpus.}
\label{tab:main_headline}
\vskip -0.2in
\end{table*}

Given the refined binary frame mask $m$, we introduce a position-wise guidance formulation to guide the logit at each denoising step. We first consider an intuitive baseline framework, termed \textbf{Swap}, which assigns the phrase-conditioned logit $c_{\text{phr}}$ inside the masked acoustic regions ($m[a]=1$) and defaults to the matrix-conditioned logit $c_{\text{src}}$ outside them ($m[a]=0$). Incorporating this routing into the standard CFG layout yields:
\begin{equation}
\tilde{l}_{\text{swap}} \;=\; \tilde c \;+\; \gamma\,(\tilde c - u),
\label{eq:swap}
\end{equation}
where $\tilde c = (1{-}m)\,c_{\text{src}} + m\,c_{\text{phr}}$. Being conceptually straightforward, \textbf{Swap} introduces native phonetics to offer accent improvements.

However, its capacity to inject a distinct foreign accent remains inherently constrained. Because the global acoustic and textual context is dominated by the matrix carrier, a more flexible control mechanism is required to amplify the language-specific guidance within the localized embedded phrase region. To rectify this architectural limitation, we algebraically decompose Eq.~\ref{eq:swap} by expanding $\tilde c$:
\begin{multline}
\tilde{l}_{\text{swap}} \;=\; \underbrace{\left[\, c_{\text{src}} + \gamma(c_{\text{src}}-u) \,\right]}_{\text{\textbf{Baseline} (Standard Global CFG)}} \\
\;+\; m\,\underbrace{(1+\gamma)}_{\text{Implicit scale}}\,(c_{\text{phr}}-c_{\text{src}}).
\label{eq:swap_expanded}
\end{multline}
Eq.~\ref{eq:swap_expanded} reveals that \textbf{Swap} is mathematically equivalent to maintaining the standard matrix-language CFG globally, while injecting a localized corrective vector pointing in the direction of $(c_{\text{phr}} - c_{\text{src}})$, which represents the direct cross-lingual contrast between the two language conditions. Crucially, however, \textbf{Swap} rigidly couples the magnitude of this contrastive injection to the global text guidance scale via the fixed coefficient $(1+\gamma)$.

To break this rigid coupling and enable customizable accent enforcement over the target acoustic segments, we propose \textbf{Phrase-Localized Language-Contrastive Guidance (LCG)}. We substitute the coupled implicit scale with an independent control parameter $\lambda$:
\begin{equation}
\tilde{l}_{\text{LCG}} \;=\; \left[\, c_{\text{src}} + \gamma\,(c_{\text{src}} - u) \,\right] \;+\; m\,\lambda\,(c_{\text{phr}} - c_{\text{src}}).
\label{eq:lcg}
\end{equation}
As in Fig.~\ref{fig:method_overview}(c), \textbf{LCG} mathematically decouples the language-steering force from the global text guidance scale $\gamma$. By routing the contrastive direction $(c_{\text{phr}} - c_{\text{src}})$ through an $m$-gated localized steering switch, \textbf{LCG} directly scales the phonetic steering intensity over the masked frames. Adjusting $\lambda$ unlocks a significantly stronger, independent guiding effect compared to the coupled \textbf{Swap} framework, driving the gated frames toward the authentic native pronunciation of the embedded phrase language without destabilizing the global matrix carrier.
\section{Experimental Setup}
\label{sec:setup}

\subsection{Datasets \& Evaluation Corpora}
\label{ssec:data}

To benchmark CS TTS, we construct a balanced, \textbf{1,200-utterance synthetic CS evaluation corpus} generated via a pipeline utilizing \texttt{gpt-5.5-2026-04-23} with xhigh reasoning effort. The corpus spans five languages (English, German, French, Japanese, Korean) perfectly balanced across twelve directional tasks (six $\text{Latin}\leftrightarrow\text{JK}$ language pairs evaluated bi-directionally, 100 utterances each). Each utterance contains 3--5 dense, technical, or literary embedded phrases (mean $\sim$3.6, $\sim$6 words, $\sim$35 characters) spanning 18 domains and 6 distinct stylistic registers. To ensure speaker continuity and isolate accent leakage from identity artifacts, each directional task is synthesized using a single fixed, high-fidelity monolingual matrix carrier reference voice prompt systematically selected from standard speech repositories.

\begin{figure}[t]
\centering
\includegraphics[width=1.0\linewidth]{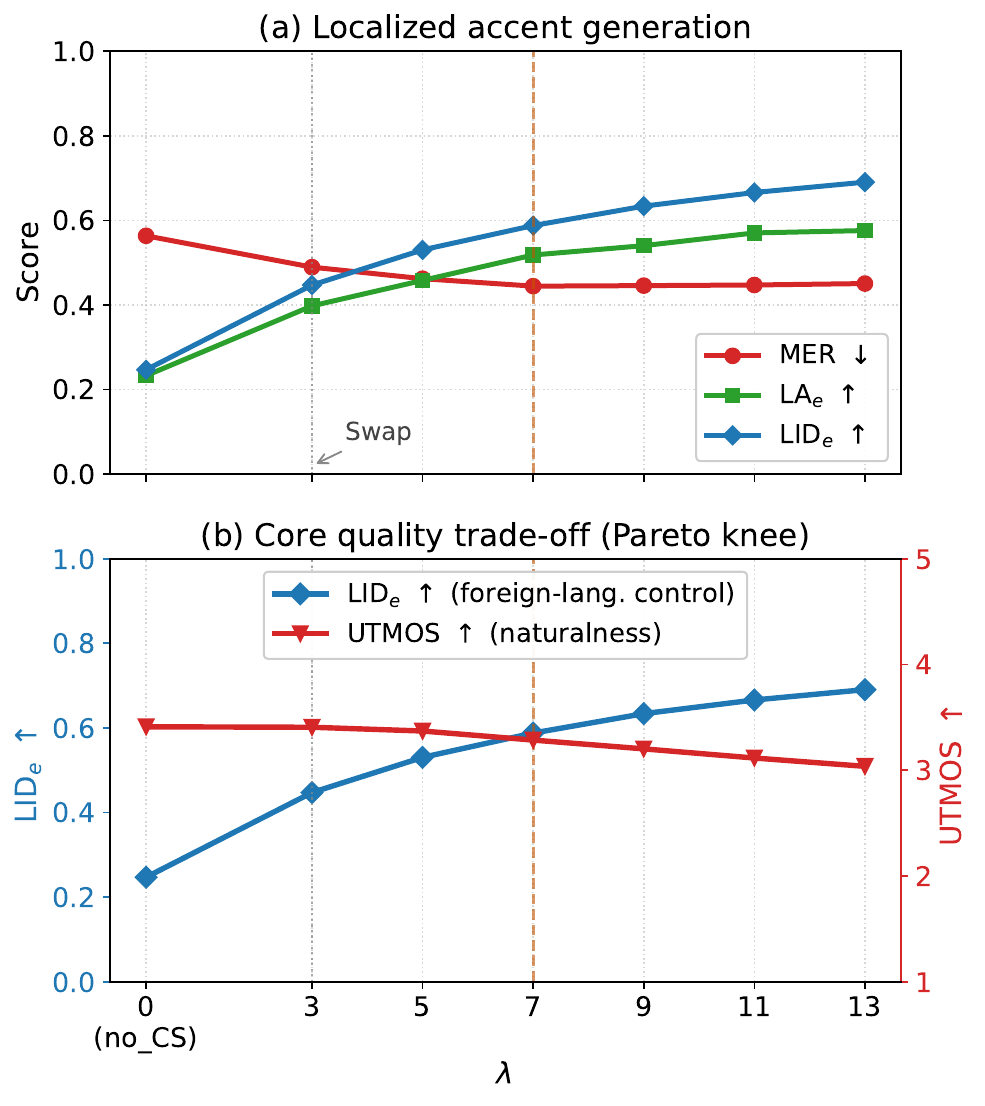}
\caption{\textbf{Localized guidance parameter sweep} on the \texttt{M4+XU} configuration. Panel (a) captures the saturation effect of the native pronunciation metrics beyond $\lambda=7$. Panel (b) details the trade-off between foreign accent enforcement ($\mathrm{LID}_e$) and global speech naturalness ($\mathrm{UTMOS}$), maximizing their Pareto envelope at $\lambda=7$.}
\label{fig:lambda_sweep}
\vskip -10pt
\end{figure}

\subsection{Comparison Systems}
\label{ssec:systems}
We evaluate our framework along three architectural axes that map to our proposed mechanics:

\paragraph{(i) Guidance Strength.} We evaluate the impact of the contrastive scaling factor $\lambda$ in Eq.~\ref{eq:lcg}. Fixing the global text guidance scale at its default value ($\gamma=2.0$), we sweep $\lambda \in \{0, 3, 5, 7, 9, 11, 13\}$ to track localized accent amplification, where $\lambda=0$ corresponds to the unguided \textbf{Baseline} (no localized accent control), while $\lambda=3$ matches the coupled \textbf{Swap} baseline (Eq.~\ref{eq:swap_expanded}).

\paragraph{(ii) Mask Refinement.} Holding guidance at $\lambda=7$, we evaluate the spatial precision-recall grid by sweeping combinations of morphological margin expansion $k \in \{0, 2, 4\}$ (effective radii $r_k \in \{0, 3, 10\}$) and the dual-tag logical union $\in \{\text{off}, \text{on}\}$ (Sec.~\ref{ssec:refine}).

\paragraph{(iii) Mask Source.} We benchmark our runtime, module-free \textbf{Probed Attention Mask} (with \texttt{M4+XU} refinement) against an offline, oracle \textbf{ForcedAligner-derived Mask} extracted via an auxiliary \texttt{Qwen3-ForcedAligner-0.6B}~\citep{Qwen3-ASR}.

\subsection{Evaluation Metrics}
\label{ssec:metrics}
Acoustic segments are indexed via Qwen3-ForcedAligner strictly for region-aware evaluation, where subscripts $_m$ and $_e$ denote the segmented matrix carrier and embedded phrase regions, respectively.

\paragraph{Objective Metrics.} We track alignment and speaker naturalness using: (1) Mixed Error Rate ($\mathrm{MER}_m$, $\mathrm{MER}_e$, $\mathrm{MER}$)~\citep{lyu10_interspeech,ugan2025pier} via Whisper-large-v3~\citep{pmlr-v202-radford23a} with language-forced decoding, computed as Word Error Rate (WER) for Latin scripts and Character Error Rate (CER) for Japanese and Korean embedded phrases; (2) Language Accuracy ($\mathrm{LA}_e$) and Identification Confidence ($\mathrm{LID}_e$) over isolated embedded audio clips, where $\mathrm{LA}_e$ tracks the ratio of segments where Whisper correctly transcribes the target embedded phrase language script during unconstrained decoding, while $\mathrm{LID}_e$ measures Whisper's language posterior probability; (3) Speaker Similarity via \texttt{WavLM-base-plus-sv}~\citep{DBLP:journals/jstsp/ChenWCWLCLKYXWZ22}, evaluating reference-to-output similarity ($\mathrm{SIM}$) and intra-utterance matrix-to-embedded consistency ($\mathrm{SIM}_{m\leftrightarrow e}$); and (4) Predicted utterance naturalness via UTMOS~\citep{saeki22c_interspeech}.

\paragraph{Subjective Metrics.} To validate actual human perceptual performance, we conduct crowd-sourced listening tests tracking two distinct axes: (1) \textbf{Global Quality MOS} over complete utterances to verify overall acoustic naturalness, and (2) \textbf{Phrase Nativeness AB Preference} over isolated segments to evaluate the perceived authenticity of the target foreign accent. The comprehensive crowdsourcing setup, step-by-step filtering protocols, task instructions, and deep methodological rationales are detailed in Appendix~\ref{sec:detailed_human_eval}.

\begin{table}[t]
\centering
\small
\setlength{\tabcolsep}{5pt}
\begin{tabular}{l c c c}
\toprule
\textbf{Metric} & \texttt{base} & \texttt{gt} (Oracle) & \texttt{M4+XU} (Ours) \\
\midrule
$\mathrm{MER}\downarrow$               & 0.564 & 0.472 & \textbf{0.445} \\
$\mathrm{MER}_m$ (Matrix) $\downarrow$ & 0.539 & 0.525 & \textbf{0.505} \\
$\mathrm{MER}_e$ (Embed.) $\downarrow$ & 0.548 & 0.365 & \textbf{0.329} \\ 
$\mathrm{LA}_e\uparrow$                & 0.233 & 0.459 & \textbf{0.518} \\
$\mathrm{LID}_e\uparrow$               & 0.247 & 0.526 & \textbf{0.588} \\
\midrule
UTMOS$\uparrow$                        & 3.411 & 3.333 & 3.285          \\
SIM$\uparrow$                          & 0.973 & 0.972 & 0.969          \\
\bottomrule
\end{tabular}
\caption{\textbf{Mask Source and Boundary Recall Analysis.} Comparative evaluation at $\lambda=7$ between the unguided baseline (\texttt{base}), the precision-anchored offline forced-alignment oracle (\texttt{gt}), and our method (\texttt{M4+XU}). Metrics represent the macro-average over all twelve directional language pairs.}
\label{tab:mask_source_comparison}
\vskip -0.2in
\end{table}

\section{Results}
\label{sec:results}

We evaluate our framework across four dimensions: macro headline performance (Sec.~\ref{ssec:headline}), spatial masking boundaries (Sec.~\ref{ssec:source_refine}), localized steering scaling regimes (Sec.~\ref{ssec:lambda_sweep}), and human perceptual preferences (Sec.~\ref{ssec:human_eval}). 

\subsection{Headline Comparison}
\label{ssec:headline}

Table~\ref{tab:main_headline} details performance across all twelve cross-lingual tasks. Averaged globally (\textbf{12-dir Overall}), our proposed \textbf{LCG} ($\lambda=7$) dramatically reduces the Mixed Error Rate ($\mathrm{MER}$) from $0.564$ to $0.445$ while more than doubling the embedded phrase language accuracy ($\mathrm{LA}_e = 0.233 \to 0.518$) and boosting identification confidence ($\mathrm{LID}_e = 0.247 \to 0.588$). Crucially, this localized foreign accent injection incurs negligible degradation to global speaker identity preservation ($\mathrm{SIM}$) and intra-utterance speaker consistency ($\mathrm{SIM}_{m\leftrightarrow e}$). Per-direction, the cross-lingual steering force is most pronounced in English-matrix configurations ($\text{EN}\to\text{JA/KO}$), effectively countering the \textbf{Baseline}'s tendency to forcefully assimilate foreign inserts into the dominant matrix accent.

\subsection{Spatial Localization: Mask Source and Refinement}
\label{ssec:source_refine}

We justify our spatial localization design by analyzing mask alternatives (Table~\ref{tab:mask_source_comparison}) and refinement configurations (Table~\ref{tab:ablation_refinement}) at a fixed scale of $\lambda=7$.

\paragraph{Mask Source Paradox.} As in Table~\ref{tab:mask_source_comparison}, our dynamic \textbf{LCG} ($\lambda=7$) outperforms the precision-heavy offline forced-alignment oracle (\texttt{gt}), achieving lower linguistic errors ($\mathrm{MER}=0.445$ vs. $0.472$) and superior script capture ($\mathrm{LA}_e=0.518$ vs. $0.459$). This paradox confirms our asymmetric error tolerance hypothesis (Sec.~\ref{ssec:span}). While offline forced-alignment truncates boundaries to maximize precision, it sacrifices transitional frame recall at code-switching junctions, triggering localized pronunciation collapse. Our refined attention mask achieves the optimal boundary recall balance without external model overhead.

\begin{table}[t]
\centering
\setlength{\tabcolsep}{5pt} 
\resizebox{\columnwidth}{!}{%
\begin{tabular}{l c c c c}
\toprule
Setup & $\mathrm{MER}\downarrow$ & $\mathrm{LA}_e\uparrow$ & $\mathrm{LID}_e\uparrow$ & UTMOS$\uparrow$ \\
\midrule
no\_CS (Baseline) & 0.564 & 0.233 & 0.247 & 3.411 \\
\midrule
$k=0$ (\texttt{base}) & 0.455 & 0.499 & 0.568 & 3.325 \\
$k=0$ + \texttt{XU}   & 0.452 & 0.493 & 0.570 & 3.329 \\
\addlinespace
$k=2$                 & 0.455 & 0.514 & 0.587 & 3.305 \\
$k=2$ + \texttt{XU}   & 0.447 & 0.517 & 0.582 & 3.304 \\
\addlinespace
$k=4$                 & 0.450 & 0.509 & 0.584 & 3.288 \\
$\boldsymbol{k=4+}$ \texttt{\textbf{XU}} \textbf{(Ours)} & \textbf{0.445} & \textbf{0.518} & \textbf{0.588} & \textbf{3.285} \\
\bottomrule
\end{tabular}
}
\caption{\textbf{Mask Refinement Ablation Study.} Performance comparison across varying margin expansion widths ($k \in \{0, 2, 4\}$) and the integration of the dual-tag logical union (\texttt{XU}) at a fixed guidance scale of $\lambda = 7$. Metrics represent the macro-average over all twelve directional language pairs.}
\label{tab:ablation_refinement}
\vskip -0.15in
\end{table}

\paragraph{Refinement Ablation.} Table~\ref{tab:ablation_refinement} tracks the spatial optimization trajectory. Progressing from a raw argmax mask ($k=0$) to our dilated setup ($k=4$ + \texttt{XU}) minimizes linguistic tracking errors ($\mathrm{MER} = 0.455 \to 0.445$). Integrating the dual-tag logical union (\texttt{XU}) acts as a zero-overhead boundary stabilizer that accommodates text-alignment track shifts across parallel tag conditions, cementing \texttt{M4+XU} as our optimal configuration. Notably, performance is largely insensitive to the exact radius: widening $r_k$ from $3$ to $10$ moves MER by only $0.002$ and $\mathrm{LID}_e$ by $0.006$, which directly supports our claim that the iterative denoising process absorbs over-dilation rather than propagating it into the matrix carrier.

\subsection{Localized Guidance Intensity ($\lambda$-Sweep)}
\label{ssec:lambda_sweep}

We analyze the impact of the contrastive scaling factor by sweeping $\lambda \in \{0, 3, 5, 7, 9, 11, 13\}$ on the \texttt{M4+XU} gating foundation (Fig.~\ref{fig:lambda_sweep}). As mapped in (a), foreign accent metrics ($\mathrm{LA}_e, \mathrm{LID}_e$) show a distinct logarithmic saturation trend, rising sharply up to $\lambda=7$ before flattening. Conversely, (b) reveals that predicted speech naturalness ($\mathrm{UTMOS}$) undergoes linear degradation as $\lambda$ escalates, with a sharp drop-off past $\lambda=7$. Notably, setting $\lambda=3$ mathematically replicates the coupled \textbf{Swap} baseline, confirming that decoupled scaling is mandatory for unconstrained steering. The functional optimization knee sits cleanly at $\lambda=7$, maximizing foreign accent enforcement while preserving global acoustic naturalness within acceptable thresholds, which is further validated as highly manageable through our informal listening tests.

\subsection{Human Perceptual Evaluation}
\label{ssec:human_eval}

We conduct two human listening tests via Amazon Mechanical Turk over $\text{JA}\to\text{EN}$ and $\text{KO}\to\text{EN}$ directions to verify our automated proxies with verified native English validators (Table~\ref{tab:human_results}), where detailed results are shown in Appendix~\ref{sec:detailed_human_eval}.

\paragraph{Phrase Nativeness Preference.} In blind forced-choice preference tests on isolated embedded acoustic segments, native speakers overwhelmingly favored our framework over the unguided \textbf{Baseline} ($\lambda=0$), yielding a decisive overall preference rate of \textbf{75.5\%} across 488 ratings (\textbf{75.7\%} for $\text{JA}\to\text{EN}$ and \textbf{75.2\%} for $\text{KO}\to\text{EN}$). This confirms that independent language-contrastive scaling successfully delivers authentic native pronunciation to natives.

\begin{table}[t]
\centering
\small
\setlength{\tabcolsep}{6pt}
\begin{tabular}{lc}
\toprule
\textbf{Metric / System} & \textbf{Overall} \\
\midrule
\multicolumn{2}{l}{\textit{Global Quality MOS (5-point)}} \\ 
~~Base ($\lambda=0$) & $4.007 \pm 0.110$ \\
~~Swap ($\lambda=3$) & $3.967 \pm 0.092$ \\
~~LCG (Ours, $\lambda=7$) & $3.931 \pm 0.092$ \\
\midrule
\multicolumn{2}{l}{\textit{Phrase Nativeness AB Pref. (Ours vs. Base)}} \\ 
~~Ours wins & \textbf{292 (59.8\%)} \\
~~Base wins & 95 (19.5\%) \\
~~Tie & 101 (20.7\%) \\
\bottomrule
\end{tabular}
\caption{\textbf{Human perceptual evaluation results} for global speech quality (5-point MOS $\uparrow$) and local phrase nativeness preference (AB Test $\uparrow$). The preference rate reported in the text excludes ties.}
\label{tab:human_results}
\vskip -10pt
\end{table}

\paragraph{Global Quality MOS.} Complete utterance evaluations on a 5-point scale demonstrate a highly stable human naturalness profile ($3.931 \pm 0.092$ for Ours vs. $4.007 \pm 0.110$ for Baseline), with a numerically small drop (< 0.08), with overlapping 95\% CIs. This crucial perceptual gate clarifies that while data-driven MOS estimators (UTMOS) penalize code-switched acoustic transitions as out-of-distribution (OOD) anomalies, human ears perceive the contextual flow as natural, verifying that \textbf{LCG} safely bypasses accent leakage without global quality degradation.
\section{Conclusion}

In this paper, we introduced a training- and module-free inference framework to mitigate cross-lingual accent leakage in phrase-level code-switching. By extracting dynamic masks from self-attention layers and introducing Phrase-Localized Language-Contrastive Guidance (LCG), our method successfully decouples localized accent control from global text guidance without any model retraining. Objective and human evaluations on a balanced 1,200-utterance 12-direction benchmark confirm that \textbf{LCG} robustly enforces native accents. Crucially, our expanded masking aligns naturally with the iterative denoising process of discrete diffusion models to absorb minor boundary errors, offering a light, zero-training-cost, and scalable alternative for controllable cross-lingual speech synthesis.

\section*{Acknowledgments}
This work was supported by the National Research Foundation of Korea (NRF) grant funded by the Korea government (MSIT) [No. 2022R1A3B1077720], the BK21 FOUR program of the Education and Research Program for Future ICT Pioneers, Seoul National University in 2024, Institute of Information \& Communications Technology Planning \& Evaluation (IITP) grant funded by the Korea government (MSIT) [NO. RS-2021-II211343, Artificial Intelligence Graduate School Program (Seoul National University), NO. RS-2022-II220959], Samsung Electronics Co., Ltd. (IO231120-07949-01 and Mobile eXperience(MX) Business), and NVIDIA Academic Grant Program.
\section*{Limitations}

While our framework enables training-free, inference-time accent control for code-switching TTS, several limitations remain. 

First, because our primary objective centers on precise diagnostic evaluation rather than large-scale dataset scaling, our evaluation corpus is relatively constrained in absolute volume, comprising 1,200 utterances. Furthermore, while this benchmark covers major high-resource language pairs, its typological and linguistic coverage is not exhaustive; consequently, the cross-lingual phonological transfer behavior of LCG within deeply low-resource or structurally divergent language families remains an open question.

Second, the proposed framework is tightly coupled with discrete diffusion language model (DLM) backbones. While logit-level contrastive steering could empirically be applied to autoregressive (AR) speech generation models, whether such localized interventions would generalize robustly or conversely exacerbate autoregressive error accumulation remains unverified, thereby bounding the immediate extensibility of our method to alternative model paradigms. Its inputs are also constrained: LCG requires the character span of the embedded phrase, which is often derivable automatically from Unicode script boundaries or a language identifier for same-script mixing, but must otherwise be supplied.

Finally, our evaluation relies on synthetically authored code-switching scripts paired with monolingual reference voices. Although we implemented a two-stage pipeline to produce diverse topics and realistic stylistic registers, these synthetic text structures may still deviate from organic real-world behaviors. Appendix~\ref{app:human_text} partially addresses this by applying LCG to human-authored code-switching transcripts, where the same ordering holds as on our synthetic benchmark. The gains carry over to the single-word insertions that dominate natural code-switching, albeit with a narrower margin and noisier phrase-level measurements over such short segments. We therefore expect the method to remain effective on real-world inputs, while its stability across the full diversity of spontaneous conversational code-switching may vary. Relatedly, such speech is often accented toward the matrix language, so native-like rendering can be viewed as a design target rather than a universal ideal; in our framework the degree of nativeness remains a per-application choice, as $\lambda$ is continuous and $\lambda = 0$ recovers the baseline.

\section*{Ethical Considerations}
While Phrase-Localized LCG significantly enhances localized accent control for code-switching speech synthesis, we acknowledge the potential ethical implications regarding voice misuse and cultural representation. The capacity for precise, training-free foreign accent steering could theoretically be exploited by malicious actors to create highly deceptive multi-lingual deepfakes. Additionally, aggressive scaling of the contrastive vector ($\lambda$) can inadvertently generate over-exaggerated phonetic patterns or stereotypical accent caricatures that risk causing cultural insensitivity.
\bibliography{custom}
\appendix
\clearpage
\section{Additional Details on Human Evaluation}
\label{sec:detailed_human_eval}

\subsection{Crowdsourcing Setup and Evaluation Protocols}
\label{app:protocol}

\begin{figure*}[p]
\centering
\fbox{\includegraphics[width=0.98\linewidth]{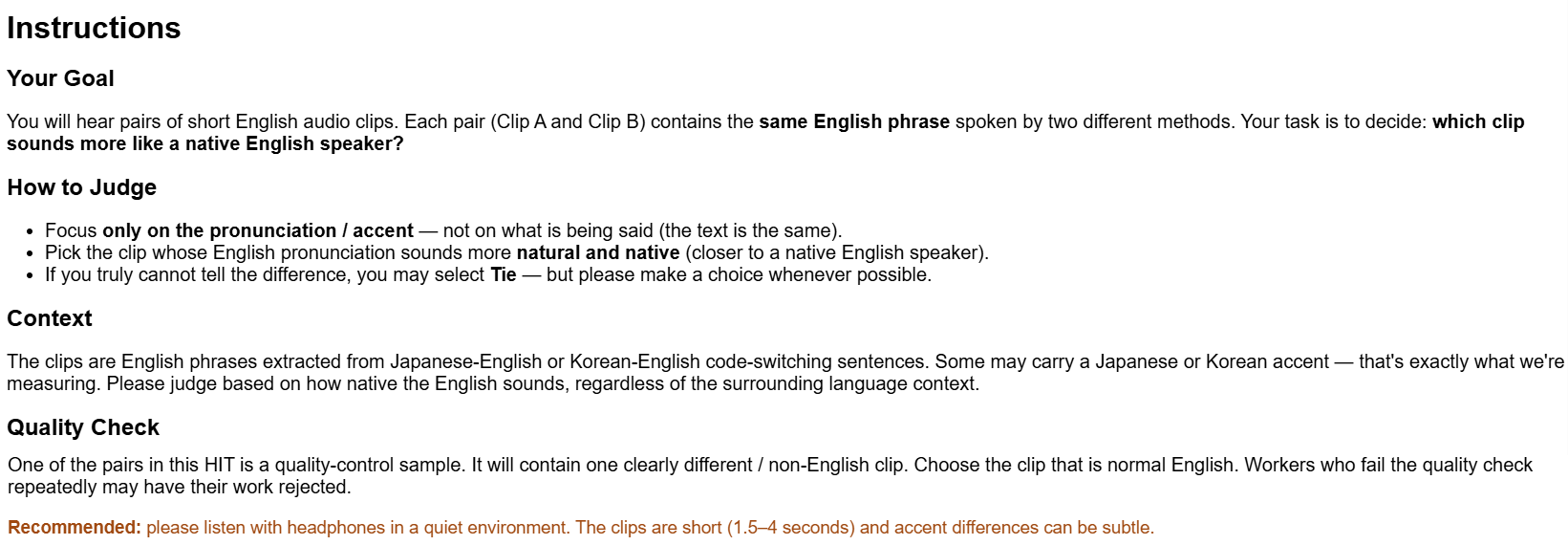}}
\vspace{1.5em}
\fbox{\includegraphics[width=0.98\linewidth]{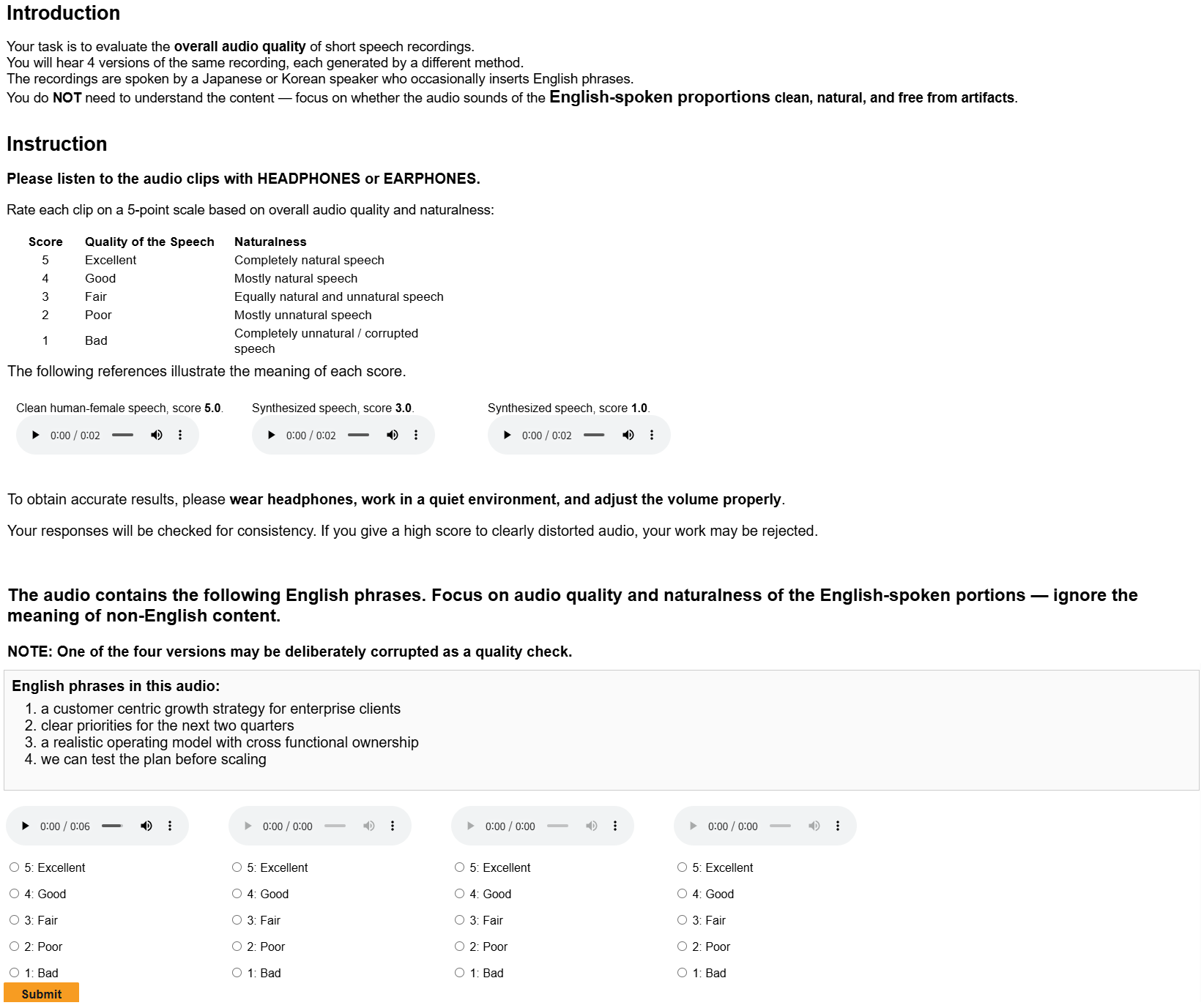}}
\caption{\textbf{Worker-facing instructions for the two human evaluation tasks.} (Top) \textbf{Phrase Nativeness AB test}: raters compare two isolated segments of the same English phrase and choose which sounds more native, with an explicit instruction to judge pronunciation rather than content. (Bottom) \textbf{Global Quality MOS test}: raters hear four full-utterance renditions of the same code-switched prompt and score each on a 5-point scale, with three labeled reference clips provided to anchor the scale. Each task embeds its own quality check: a pair containing one clearly non-English clip in the AB test, and a deliberately corrupted rendition among the four versions in the MOS test.}
\label{fig:amt}
\end{figure*}

All human perceptual annotations were crowd-sourced via the Amazon Mechanical Turk (AMT) platform, strictly localized to native English-proficient geographic regions. Individual tasks were purposefully structured to minimize rater fatigue, utilizing a single-stage validation filter executed at the individual HIT level: any submission containing an incorrect answer on the embedded quality-check item was systematically discarded from our final post-filtered analysis. Given that the AMT workforce predominantly consists of native English speakers, our human evaluation was selectively conducted on English-inclusive code-switched utterances. The complete worker-facing instructions for both tasks are shown in Fig.~\ref{fig:amt}.

\paragraph{Global Quality MOS Protocol.} Raters were directed to listen to all four audio samples associated with a given code-switched text prompt (representing Baseline, Swap, LCG, and a low-quality anchor) and rate each file on a 5-point scale (\emph{Excellent / Good / Fair / Poor / Bad}). The judgment criteria mandated prioritizing overall speech naturalness, prosodic continuity, and the absence of processing anomalies. A total of $N=275$ valid full-utterance HITs across 199 unique native English workers were secured after filtering.

\paragraph{Phrase Nativeness AB Protocol.} To strictly isolate phonetic accent steering from token generation failures, the comparison pairs were curated exclusively from the oracle subset where all candidate configurations achieved perfect language accuracy ($\mathrm{LA}_e=1$). Workers were presented with a target English text phrase and two short isolated audio segments (A and B) both intended to express that phrase. Raters answered the forced-choice question: \emph{``Which clip sounds more like a native English speaker?''} utilizing three response options (\emph{Clip A / Clip B / Tie}). Instructions clarified that both segments share identical lexical text, isolating the judgment to accent authenticity rather than content. We secure $122$ valid HITs from 101 unique workers, translating to $488$ total pair-level preference ratings.

\paragraph{Worker Compensation and Research Disclosure.} 
The introductory dashboard explicitly disclosed to all participants that the audio assets were synthetic samples generated for a cross-lingual code-switching text-to-speech research study, and that the collected annotations would be utilized strictly to evaluate perceptual quality and accent nativeness. Each assignment was compensated at a rate of \$0.10--\$0.20 per HIT based on task complexity, where an expected completion time is 30--120 seconds per task. Participation in any HIT was entirely voluntary, and no demographic attributes or personally identifiable information (PII) were collected at any stage of the study.

\begin{table*}[t]
\centering
\small
\begin{tabular}{lcccc}
\toprule
\textbf{Direction} & \textbf{System Configuration} & \textbf{Valid HITs ($N$)} & \textbf{Quality MOS} & \textbf{95\% CI} \\
\midrule
\multirow{3}{*}{$\text{JA}\to\text{EN}$ (Raters = 111)}
& Vanilla CFG ($\lambda=0$) & 136 & 3.787 & $\pm$0.162 \\
& Swap Baseline ($\lambda=3$)  & 136 & 3.728 & $\pm$0.130 \\
& LCG (Ours, $\lambda=7$)      & 136 & 3.757 & $\pm$0.142 \\
\midrule
\multirow{3}{*}{$\text{KO}\to\text{EN}$ (Raters = 116)}       
& Vanilla CFG ($\lambda=0$) & 139 & 4.223 & $\pm$0.140 \\
& Swap Baseline ($\lambda=3$)  & 139 & 4.201 & $\pm$0.119 \\
& LCG (Ours, $\lambda=7$)      & 139 & 4.101 & $\pm$0.112 \\
\midrule
\multirow{3}{*}{Overall (Raters = 199)}       
& Vanilla CFG ($\lambda=0$) & 275 & 4.007 & $\pm$0.110 \\
& Swap Baseline ($\lambda=3$)  & 275 & 3.967 & $\pm$0.092 \\
& LCG (Ours, $\lambda=7$)      & 275 & 3.931 & $\pm$0.092 \\
\bottomrule                                            
\end{tabular}
\caption{\textbf{Detailed Quality MOS results per language direction} after quality-check item filtering. All three code-switching execution patterns are statistically indistinguishable due to overlapping 95\% confidence intervals, demonstrating that LCG preserves global carrier naturalness.}
\label{tab:mos_per_direction}
\end{table*}

\begin{table*}[t]
\centering
\small
\begin{tabular}{lccccc}
\toprule
\textbf{Direction} & \textbf{Total Pairs ($N$)} & \textbf{Ours Wins} & \textbf{Vanilla Wins} & \textbf{Tie} & \textbf{Ours Pref. Rate (excl. Tie)} \\                                
\midrule
$\text{JA}\to\text{EN}$ & 237 & 134 & 43 & 60 & \textbf{75.7\%} \\
$\text{KO}\to\text{EN}$ & 251 & 158 & 52 & 41 & \textbf{75.2\%} \\
\midrule
Overall                 & 488 & 292 & 95 & 101 & \textbf{75.5\%} \\                                    
\bottomrule
\end{tabular}
\caption{\textbf{Detailed Phrase Nativeness AB Test results} evaluated over the oracle $\mathrm{LA}_e=1$ phrase subset. Our proposed localized guidance framework is preferred over the unguided baseline by a substantial margin across all language configurations ($p<0.001$).}
\label{tab:ab_per_direction}
\end{table*}

\subsection{Methodological Rationale for the Two-Track Design}
\label{app:eval_rationale}

Evaluating code-switching (CS) speech synthesis via a single global naturalness metric often conflates distinct perceptual dimensions. To resolve this, we implement the decoupled, two-track subjective evaluation protocol detailed above, driven by two primary methodological requirements:

\paragraph{(i) Quality MOS supplements an unreliable automatic proxy.}
Automatic naturalness estimators such as UTMOS~\citep{saeki22c_interspeech} are trained predominantly on monolingual, single-language speech distributions. Consequently, they tend to treat the dense, localized acoustic and phonetic transitions characteristic of intra-utterance code-switching as out-of-distribution (OOD) artifacts, frequently penalizing well-rendered foreign phrases as synthesis anomalies. Conducting a complete-utterance MOS with native listeners allows us to cross-verify that scaling up our localized accent control vector ($\lambda=7$) does not inadvertently compromise global prosodic flow, transition smoothness, or carrier speaker identity.

\paragraph{(ii) AB Test on the $\mathrm{LA}_e=1$ subset isolates the perceptual signal of accent control.} 
An absolute category rating (MOS) on isolated phrase segments inherently conflates two orthogonal axes: \emph{generation stability} (i.e., whether the system successfully synthesized intelligible speech tokens) and \emph{phonetic nativeness} (i.e., whether the realized pronunciation sounds authentic to native ears). By pre-screening our evaluation set to an oracle subset where all comparison systems achieve perfect language accuracy ($\mathrm{LA}_e=1$), we effectively eliminate the generation success variance. Forced-choice AB testing over this cleansed distribution then probes \emph{exclusively} the residual accent-nativeness preference, isolating the precise phonetic steering force that LCG is engineered to modulate.

\subsection{Quality MOS Test: Per-Direction Analysis}
\label{app:mos_details}

The comprehensive per-direction results for full-utterance naturalness are detailed in Table~\ref{tab:mos_per_direction}. Crucially, a directional breakdown confirms that across both the $\text{JA}\to\text{EN}$ and $\text{KO}\to\text{EN}$ tasks, the global speech naturalness scores of our proposed configuration (\texttt{M4+XU}, $\lambda=7$) remain statistically indistinguishable from both the unguided baseline ($\lambda=0$) and the coupled Swap ($\lambda=3$). The overlapping 95\% confidence intervals across all evaluation branches empirically validate our spatial isolation thesis: restricting the language-contrastive steering force to dynamically probed attention boundaries successfully insulates the unswitched matrix regions from acoustic degradation.

\subsection{Phrase Nativeness AB Test: Per-Direction Analysis}
\label{app:ab_details}

Table~\ref{tab:ab_per_direction} outlines the directional performance profiles for localized phrase nativeness. The empirical results demonstrate that regardless of the source matrix language direction, our framework is overwhelmingly preferred by native validators for rendering more authentic and native-like phrasal pronunciations. Specifically, our configuration secures a decisive preference rate of \textbf{75.7\%} in the $\text{JA}\to\text{EN}$ track and \textbf{75.2\%} in the $\text{KO}\to\text{EN}$ track, culminating in a robust \textbf{75.5\%} overall preference margin. This language-agnostic consistency is highly statistically significant ($z=10.01$ against a chance performance of 50\%, $p<0.001$), proving that LCG successfully overrides host-language accent assimilation across structurally distinct typological environments.

\section{Implementation Details}
\label{ssec:impl}

Our framework is built on the public 0.81\,B-parameter \texttt{k2-fsa/OmniVoice} discrete DLM backbone (28 decoder layers, 8-codebook codec)~\citep{zhu2026omnivoice}. Sampling executes over $T=32$ iterative denoising steps under a default CFG scale of $\gamma=2$. Our core LCG configuration operates at $\lambda=7$ with \texttt{M4+XU} refinement, selected as the optimal operational knee that balances aggressive localized language steering and global acoustic/speaker stability. All other parameters mirror the baseline OmniVoice environment.

\section{Runtime Cost of Attention Extraction}
\label{app:backend}

LCG operates by reading an internal signal, the decoder's self-attention weights, during the forward pass. This imposes one practical constraint at deployment: fused attention kernels such as FlashAttention and SDPA never materialize the full attention matrix and therefore do not return attention weights. Recovering them requires either recomputing attention outside the fused kernel or falling back to an eager implementation, and our mask extraction takes the latter route. All guidance conditions compared in this paper are run on the same eager backend, so the reported contrasts are backend-fair. We note that the OmniVoice reference implementation runs on SDPA rather than FlashAttention, so we benchmark against SDPA.

To quantify the deployment cost, we time $50$ benchmark utterances ($25$ JA\,$\rightarrow$\,EN and $25$ KO\,$\rightarrow$\,EN; the identical utterances across all conditions, mean duration $37.7$\,s) on a single NVIDIA B200 GPU in bfloat16, using the decoding configuration of the main experiments ($T = 32$ steps, $\gamma = 2$, and $\lambda = 7$ with \texttt{M4+XU} for LCG). Two warmup utterances per condition are excluded from the timing. We report the real-time factor (RTF), the wall-clock synthesis time divided by the duration of the generated audio, together with the absolute synthesis time per utterance. Table~\ref{tab:backend} summarizes the results.

\begin{table}[t]
\centering
\small
\setlength{\tabcolsep}{4pt}
\begin{tabular}{llccc}
\toprule
Condition & Backend & RTF $\downarrow$ & Time (s) & vs.\ SDPA \\
\midrule
Baseline            & SDPA  & 0.024 & 0.89 & 1.00$\times$ \\
Baseline            & eager & 0.038 & 1.43 & 1.61$\times$ \\
LCG ($\lambda{=}7$) & eager & 0.053 & 2.01 & 2.26$\times$ \\
\bottomrule
\end{tabular}
\caption{Synthesis cost by attention backend, over $50$ utterances of $\sim$$38$\,s audio. ``Time'' is the mean wall-clock synthesis time per utterance, excluding model load and I/O; ``vs.\ SDPA'' is its ratio to the SDPA baseline, computed from unrounded measurements. LCG requires the eager backend because it consumes attention weights.}
\label{tab:backend}
\end{table}

The total $2.26\times$ overhead decomposes into $1.61\times$ from the eager backend and $1.40\times$ from the third guidance branch. The latter sits below the $1.5\times$ FLOPs that one additional branch would imply, indicating that mask extraction itself introduces minimal computational burden. In absolute terms LCG still synthesizes about $19\times$ faster than real time, costing $+1.1$\,s per $\sim$$38$\,s utterance. The eager cost is also reducible in principle: only $2$ of the $28$ decoder layers are probed, so fused kernels could be retained for the remaining $26$.

\section{Robustness to the Attention Weighting}
\label{app:vnorm}

LCG derives its phrase mask from where each acoustic frame attends within the transcript, so its correctness rests on attention being a faithful alignment signal. A line of work on model interpretability cautions against reading attention this way: raw attention weights can overstate a token's contribution when the corresponding value vectors are small in norm \citep{kobayashi-etal-2020-attention}, and attention sinks can further concentrate weight on positions that carry little content.

Three properties of our setup limit this concern. (i) We use attention as a text-to-audio \emph{alignment} signal, following its classical role in TTS and ASR alignment, rather than as a general measure of token attribution. (ii) The argmax is restricted to transcript token columns, so typical sink positions such as sequence-initial control tokens are excluded by construction. (iii) The layer/head configuration was not chosen a priori but selected empirically by recall against forced-alignment ground truth, as shown in Sec.~\ref{ssec:span}.

Rather than rest on these properties alone, we test the concern directly. We re-scored twelve layer/head configurations from our sweep, including the top-ranked one, under the value-norm weighting $\alpha \cdot \lVert W_O v \rVert$ of \citet{kobayashi-etal-2020-attention}, computed per head before any head or layer pooling. Scoring follows the Fig.~\ref{fig:attention_probe} setup: KSS and LJSpeech pilot sets, $N = 100$ utterances each, with masks scored against forced-alignment ground truth.

\begin{table}[t]
\centering
\small
\setlength{\tabcolsep}{4pt}
\begin{tabular}{lcccl}
\toprule
& \multicolumn{3}{c}{Phrase recall $\uparrow$} & \\
\cmidrule(lr){2-4}
Weighting & en & ko & worst & Rank-1 \\
\midrule
raw $\alpha$ (ours)              & 0.407 & 0.467 & 0.407 & $\{8,12\}$, max \\
$\alpha \lVert W_O v \rVert$     & 0.408 & 0.469 & 0.408 & $\{8,12\}$, max \\
\bottomrule
\end{tabular}
\caption{Phrase-mask recall under raw versus value-norm-weighted attention, on the English (LJSpeech) and Korean (KSS) pilot sets. ``worst'' is the worst-case recall across the two domains, the criterion used for configuration selection; ``Rank-1'' is the configuration that criterion selects among the twelve re-scored comparators.}
\label{tab:vnorm}
\end{table}

Two observations follow from Table~\ref{tab:vnorm}. First, the selection is unchanged: $\{L_8, L_{12}\}$ with head-max pooling remains rank-1 in worst-case recall across this comparator set under both weightings. Second, recall at our operating point is nearly identical under the two weightings, so raw attention costs no recall in this setting, and the residual difference is far smaller than the boundary tolerance that the margin dilation already provides.

\section{Evaluation on Human-Authored Code-Switching Text}
\label{app:human_text}

\begin{table*}[t]
\centering
\small
\begin{tabular}{llcccccc}
\toprule
Regime & Condition & MER $\downarrow$ & $\mathrm{MER}_e \downarrow$ &
$\mathrm{LA}_e \uparrow$ (95\% CI) & $\mathrm{LID}_e \uparrow$ &
UTMOS $\uparrow$ & SIM $\uparrow$ \\
\midrule
\multirow{3}{*}{Long-phrasal ($\ge 4$-word)}
 & Baseline ($\lambda{=}0$)        & 0.492 & 0.284 & 0.196 $\pm$ 0.074 & 0.158 & 3.84 & 0.970 \\
 & Swap ($\lambda{=}3$)            & 0.477 & 0.257 & 0.290 $\pm$ 0.085 & 0.217 & 3.87 & 0.969 \\
 & LCG ($\lambda{=}7$, ours)       & \textbf{0.433} & \textbf{0.232} & \textbf{0.358} $\pm$ 0.090 & \textbf{0.311} & 3.87 & 0.969 \\
\midrule
\multirow{3}{*}{Single-word}
 & Baseline ($\lambda{=}0$)        & 0.201 & 0.898 & 0.217 $\pm$ 0.079 & 0.140 & 3.76 & 0.969 \\
 & Swap ($\lambda{=}3$)            & 0.197 & 0.980 & 0.302 $\pm$ 0.089 & 0.198 & 3.77 & 0.968 \\
 & LCG ($\lambda{=}7$, ours)       & 0.203 & 0.943 & \textbf{0.363} $\pm$ 0.093 & \textbf{0.225} & 3.76 & 0.969 \\
\midrule
\multirow{3}{*}{All ($N = 200$)}
 & Baseline ($\lambda{=}0$)        & 0.346 & 0.591 & 0.206 $\pm$ 0.054 & 0.149 & 3.80 & 0.970 \\
 & Swap ($\lambda{=}3$)            & 0.337 & 0.618 & 0.296 $\pm$ 0.061 & 0.208 & 3.82 & 0.969 \\
 & LCG ($\lambda{=}7$, ours)       & \textbf{0.318} & \textbf{0.587} & \textbf{0.360} $\pm$ 0.065 & \textbf{0.268} & 3.81 & 0.969 \\
\bottomrule
\end{tabular}
\caption{Evaluation on human-authored Korean--English code-switching transcripts, 100 utterances per stratum. The ordering Baseline $<$ Swap $<$ LCG holds on both embedded-phrase nativeness metrics ($\mathrm{LA}_e$ and $\mathrm{LID}_e$) in both strata, while UTMOS and speaker similarity remain flat. One single-word candidate was dropped during sampling: its \emph{baseline} rendition drives the ASR into a repetition loop ($\mathrm{MER}_e = 48$ against a one-word reference), which alone would raise that stratum's baseline $\mathrm{MER}_e$ by $\approx 0.47$. The exclusion is conservative, since it improves the baseline, and it affects only $\mathrm{MER}_e$.}
\label{tab:human_text}
\end{table*}

Our evaluation in the main experiments uses a synthetic benchmark: a mixed-language \emph{text} corpus, with the speech under evaluation always synthesized by the systems being compared. We chose synthetic text because it lets us control the factors that matter for diagnosing a model's code-switching behavior: dense multi-word phrasal insertions, balanced coverage of all twelve directions, controlled phrase density and position, and a fixed reference voice per direction. To verify that the effect is not an artifact of that choice and that it carries over to real-world inputs, we repeat the evaluation on human-authored code-switching transcripts.

We use Korean--English mixed-speech transcripts of real conversations\footnote{We used datasets (한영 혼합 인식 데이터, Korean--English mixed-speech recognition data) from The Open AI Dataset Project (AI-Hub, S. Korea). All data information can be accessed through `AI-Hub (www.aihub.or.kr)'.}, from which we sample two strata of $100$ utterances: \emph{long-phrasal} ($\ge 4$-word English insertions, matching our benchmark's dense-insertion regime) and \emph{single-word} (the regime that dominates natural code-switching). Only the transcripts are used; no recorded audio enters the pipeline. All other settings match the main paper: the same backbone, the ko\,$\rightarrow$\,en reference voice, $\gamma = 2$, \texttt{M4+XU}, and $\lambda \in \{0, 3, 7\}$, all scored with the identical evaluation pipeline.

The trend replicates on real-world text (Table~\ref{tab:human_text}): the ordering is identical (Baseline $<$ Swap $<$ LCG) on both embedded-phrase nativeness metrics ($\mathrm{LA}_e$ and $\mathrm{LID}_e$) in both strata, with LCG raising embedded-phrase language accuracy over the baseline by $+0.154$ overall, while UTMOS stays flat and speaker similarity is unchanged. The gain is largest in the long-phrasal regime, consistent with our claim that dense multi-word insertions are where accent leakage is most severe. The single-word regime, which occurs most often in natural code-switching, remains noisier: its sub-second segments make language identification unreliable and $\mathrm{MER}_e$ unstable, but LCG still improves embedded-phrase accuracy there, if by a smaller margin.

\begin{table}[t]
\centering
\small
\setlength{\tabcolsep}{4pt}
\begin{tabular}{lccc}
\toprule
System (12-direction overall) & MER $\downarrow$ & $\mathrm{LA}_e \uparrow$ & $\mathrm{LID}_e \uparrow$ \\
\midrule
MOSS-TTS-v1.5 (8B)              & 0.584 & 0.280 & 0.302 \\
OmniVoice baseline (0.81B)      & 0.564 & 0.233 & 0.247 \\
\;\; + LCG ($\lambda{=}7$, ours) & \textbf{0.445} & \textbf{0.518} & \textbf{0.588} \\
\bottomrule
\end{tabular}
\caption{System-level reference comparison on the full $1{,}200$-utterance benchmark. Note that this result is a cross-backbone reference point, not a mechanism-isolating comparison.}
\label{tab:external}
\end{table}

\begin{table*}[t]
\small
\begin{center}
\begin{tabular}{l|l|l}
\toprule
\textbf{Category} & \textbf{Name (with Link)} & \textbf{License} \\ \midrule
\multirow{6}{*}{Model}  
& \href{https://github.com/k2-fsa/OmniVoice}{OmniVoice} \cite{zhu2026omnivoice} & Apache-2.0 \\
& \href{https://huggingface.co/openai/whisper-large-v3}{Whisper-large-v3} \cite{pmlr-v202-radford23a} & Apache-2.0 \\
& \href{https://huggingface.co/microsoft/wavlm-base-plus-sv}{WavLM-base-plus-sv} \cite{DBLP:journals/jstsp/ChenWCWLCLKYXWZ22} & CC BY-SA 3.0 \\
& \href{https://github.com/sarulab-speech/UTMOS22}{UTMOS} \cite{saeki22c_interspeech} & MIT \\
& \href{https://huggingface.co/Qwen/Qwen3-ForcedAligner-0.6B/}{Qwen3-ForcedAligner-0.6B} \cite{Qwen3-ASR} & Apache-2.0 \\ 
& \href{https://huggingface.co/OpenMOSS-Team/MOSS-TTS-v1.5}{MOSS-TTS-v1.5} \cite{gong2026mossttstechnicalreport} & Apache-2.0 \\ \midrule
\multirow{4}{*}{Dataset} 
& \href{https://keithito.com/LJ-Speech-Dataset/}{LJSpeech} \cite{ljspeech17} & Public Domain \\
& \href{https://huggingface.co/datasets/Bingsu/KSS_Dataset}{KSS Dataset} \cite{park2018kss} & CC BY-NC-SA 4.0 \\
& \href{https://huggingface.co/datasets/ylacombe/cml-tts}{CML-TTS} \cite{Cmltts2023} & CC BY 4.0 \\
& \href{https://github.com/Kyubyong/css10}{CSS10} \cite{park2019css10} & Apache-2.0 \\
\bottomrule
\end{tabular}
\vskip -0.05in
\caption{Software licenses and hyperlinked references of the models and datasets utilized in this work.}
\vskip -0.2in
\label{table:resource_licenses_compact}
\end{center}
\end{table*}

\section{Reference Comparison with an External System}
\label{app:external}

To compare the effect of our training-free, inference-time guidance against an external TTS system, we additionally evaluate MOSS-TTS-v1.5 (8B)~\citep{gong2026mossttstechnicalreport}, a recently released model that documents code-switching as a supported capability and covers all five of our languages. We synthesize all $1{,}200$ benchmark utterances with the official checkpoint, using zero-shot cloning from the same per-direction reference voices, the matrix language passed as its single utterance-level language parameter (the same conditioning granularity as our unguided baseline), and the model-card default decoding settings. Scoring uses our identical evaluation pipeline.

Despite being roughly $10\times$ larger and labeled with code-switching as a capability, MOSS-TTS lands near our unguided baseline on embedded-phrase nativeness ($\mathrm{LA}_e$ $0.280$ vs.\ $0.233$; $\mathrm{LID}_e$ $0.302$ vs.\ $0.247$), suggesting that global utterance-level language conditioning does not by itself resolve accent leakage regardless of model scale. Its performance is also highly uneven across directions: $\mathrm{LA}_e$ reaches $0.59$ to $0.88$ on the four English--Japanese/Korean directions, approaching our $\lambda = 7$ configuration on EN\,$\rightarrow$\,KO ($0.884$ vs.\ $0.887$), but stays below $0.21$ on the eight directions involving German or French, at or below the level of our unguided baseline. LCG, by contrast, raises nativeness consistently across all twelve directions on the same backbone. As the two systems differ in backbone and training data, we read this as a reference point rather than a controlled comparison.

\section{Licenses}
\label{sec:licenses}

Table~\ref{table:resource_licenses_compact} outlines the licensing specifications for datasets and models leveraged in this work. The Korean--English mixed-speech transcripts used in Appendix~\ref{app:human_text} are additionally used under a research-use agreement; we therefore report aggregate results only. Every other asset is publicly accessible for research purposes and was utilized strictly within academic boundaries. Furthermore, each component dataset was rigorously curated for research applications and contains no personally identifiable information (PII) or offensive content.

\section{Usage of AI Assistant}
\label{sec:usage}
This paper is written with the help of AI assistant, Gemini, ChatGPT, and Claude. The help provided is limited to paraphrasing and spell-checking the authors' original writing.

\end{document}